%% file: main.tex
\documentclass[10pt,twocolumn,letterpaper]{article}

\usepackage{cvpr}              


\definecolor{cvprblue}{rgb}{0.21,0.49,0.74}
\usepackage[pagebackref,breaklinks,colorlinks,allcolors=cvprblue]{hyperref}

\def\paperID{17} 
\def\confName{CVPRW}
\def\confYear{2025}

\definecolor{somegray}{rgb}{0.5, 0.5, 0.5}
\newcommand{\darkgrayed}[1]{\textcolor{somegray}{#1}}
\makeatletter
\newcommand*\titleheader[1]{\gdef\@titleheader{#1}}
\AtBeginDocument{%
  \let\st@red@title\@title
  \def\@title{%
    \vskip-3em
    \bgroup\normalfont\large\centering\@titleheader\par\egroup
    \vskip1.5em\st@red@title}
}
\makeatother
\titleheader{\darkgrayed{This paper has been accepted for publication at the\\
IEEE Conference on Computer Vision and Pattern Recognition Workshops (CVPRW), Nashville, 2025.
\copyright IEEE}}
\title{ForesightNav: Learning Scene Imagination for Efficient Exploration}
\author{
\and
\and
Hardik Shah\\
ETH Zurich
\and
Jiaxu Xing\\
University of Zurich
\and
Nico Messikommer\\
University of Zurich
\and
Boyang Sun\\
ETH Zurich
\and
\and
\and
Marc Pollefeys\\
ETH Zurich, Microsoft
\and
Davide Scaramuzza\\
University of Zurich\\
{}
}

\begin{document}
\maketitle
\input{sec/0_abstract}    
\input{sec/1_intro}
\input{sec/2_related_work}
\input{sec/3_Approach}
\input{sec/4_Experiments}

\input{sec/5_Conclusion}
\input{sec/6_Acknowledgement}
{
    \small
    \bibliographystyle{ieeenat_fullname}
    \bibliography{main}
}

\input{sec/X_suppl}

\end{document}

%% file: sec/0_abstract.tex
\begin{abstract}
Understanding how humans leverage prior knowledge to navigate unseen environments while making exploratory decisions is essential for developing autonomous robots with similar abilities.
In this work, we propose \textbf{ForesightNav}, a novel exploration strategy inspired by human imagination and reasoning. 
Our approach equips robotic agents with the capability to predict contextual information, such as occupancy and semantic details, for unexplored regions.
These predictions enable the robot to efficiently select meaningful long-term navigation goals, significantly enhancing exploration in unseen environments.
We validate our imagination-based approach using the Structured3D dataset, demonstrating accurate occupancy prediction and superior performance in anticipating unseen scene geometry. 
Our experiments show that the imagination module improves exploration efficiency in unseen environments, achieving a 100\% completion rate for PointNav and an SPL of 67\% for ObjectNav on the Structured3D Validation split.
These contributions demonstrate the power of imagination-driven reasoning for autonomous systems to enhance generalizable and efficient exploration. Project code is available at \href{https://github.com/uzh-rpg/foresight-nav}{uzh-rpg/foresight-nav}.

\begin{figure}[ht]
    \centering
    \includegraphics[width=0.475\textwidth]{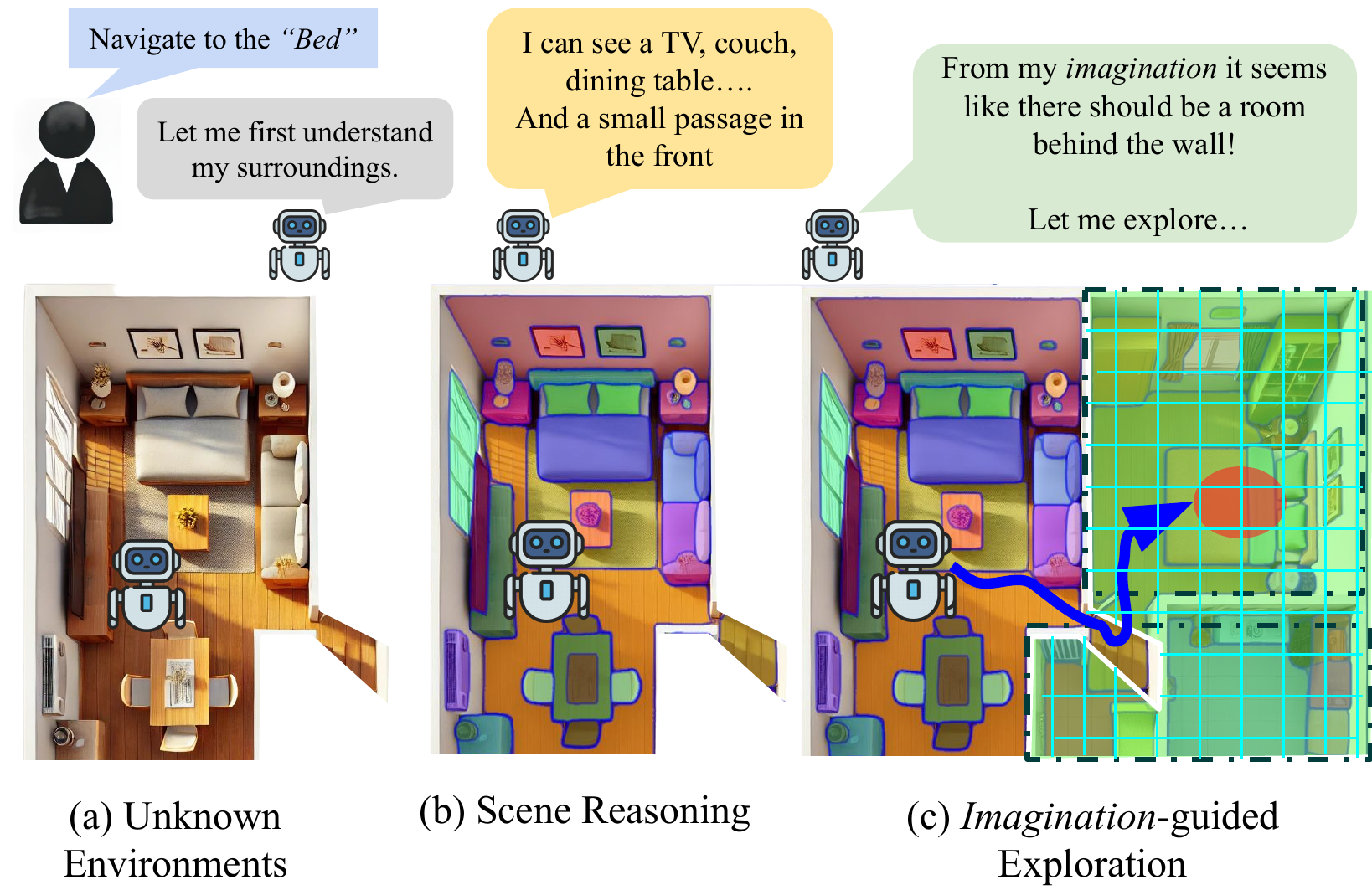}
    \caption{ForesightNav proposes \textit{Imagination} aided exploration in unknown environments. The past observations by the agent are used to reason about the scene and extract geometric and semantic information. The agent uses this information to \textit{imagine} the structure of the unobserved regions, enhancing its ability to explore efficiently.}
    \label{fig:teaser_img}
\end{figure}
\end{abstract}

%% file: sec/1_intro.tex
\section{Introduction}
\label{sec:introduction}
Human navigation in novel environments is a complex process guided by a combination of explicit maps and accumulated geometric and semantic information.
This prior information allows us to infer spatial layouts and anticipate the presence and location of objects based on context and experience.
For example, we intuitively expect to find toilets and showers in bathrooms or look for a pot in the kitchen, even without seeing them first.
Given only a partial view, humans use their \emph{imagination} to reason about unexplored areas, predicting unseen regions based on prior knowledge to guide navigation and exploration decisions~\cite{epstein2017cognitive}.
Extending this imagination capability to autonomous robots would enhance their ability to understand and interpret complex environments.
Such improved understanding directly translates to more efficient autonomous robotic navigation.
Consequently, incorporating imagination-based capabilities into robotic systems is promising for real-world applications such as home service, search and rescue, delivery, and industrial inspections~\cite{xing2023autonomous, mishra2020drone}.

However, accurately predicting unseen regions from limited or partial observations remains challenging, as it requires robust generalization to novel environments.
Moreover, developing suitable representations and training data for imagination networks that effectively include geometric structure and semantic information is inherently complex.
Existing navigation approaches have progressed by integrating scene information in an end-to-end fashion, often via reinforcement learning (RL). 
However, these end-to-end pipelines frequently fail to generalize and demand extensive retraining for each new scenario~\cite{batra2020objectnav, chaplot2020object, zeng2024poliformer}.
In contrast, advanced Vision-Language Models (VLMs) offer zero-shot solutions that circumvent task-specific training, though they may struggle with navigation precision due to their limit in understanding spatial information~\cite{qu2024ippon, zhang2024tag, rana2023sayplan}.

Very few prior works~\cite{katyal2019uncertainty, schmid2022scexplorer, ericson2024beyond, zhang2024imagine} have focused on imagining explicit map representation from partial observation, like we humans naturally do and have limitations in their scalability and evaluation.
To address this, we propose a novel exploration strategy for autonomous navigation that equips agents with a learned imagination capability, as illustrated in Figure \ref{fig:teaser_img}. 
Our method models imagination as a neural network that leverages current and past observations to predict unobserved regions of the environment, guiding more informed and goal-directed exploration.
Our approach comprises two main components: \textit{predicting scene geometry} from partially observed occupancy maps and enhancing this prediction with \textit{semantic understanding} using visual language features.
Inspired by human intuition, our approach forms beliefs about unobserved areas based on semantic relationships and spatial understanding, facilitating efficient and informed exploration.
The Imagination Module is compatible with predicting both the unseen floor plan geometry and semantic visual-language (CLIP) features~\cite{radford2021learning} from partially observed information.
This unified prediction allows the agent to efficiently select long-term navigation goals using limited partial observation.
Our results demonstrate superior performance in point goal navigation in unseen environments.
The imagination strategy significantly improves navigation efficiency by predicting unseen scene geometry from partial occupancy information. 
We outperform SOTA approaches in the Object Goal Navigation task on the Structured3D dataset that requires both semantic and geometric analysis. 
To summarize, our contributions are: 
\begin{enumerate}
    \item We propose \textbf{ForesightNav}, a novel approach for exploration based on \textbf{Imagination} of the scene representation from past observations, where an agent goes beyond reactive strategies and actively reasons about the unseen environment. 
    
    \item We propose a unique method for the training of occupancy and semantic prediction of indoor scenes by \textbf{generating partially observed occupancy masks} involving the simulation of an agent in an unseen 2D grid world of a scene to reach uniformly distributed goal waypoints.
    
    \item We introduce a new \textbf{closed-loop evaluation benchmark} for the task of Object Goal Navigation in a 2D grid world using the Structured3D dataset, with 3500 indoor scenes.
    Our benchmark on Structured3D enables a more comprehensive evaluation than existing benchmarks, improving the assessment of method generalization
\end{enumerate}

%% file: sec/2_related_work.tex
\section{Related Work}
\label{sec:related_work}

\subsection{Object Goal Navigation}

Object Goal Navigation (ObjectNav) focuses on navigating to a specified object in a novel environment, utilizing semantic priors to improve navigation efficiency. 
Traditional approaches leverage reinforcement learning~\cite{ye2021auxiliary, chang2020semantic, procthor,gireesh2022object}, learning from demonstration~\cite{ramrakhya2023pirlnav}, or creating semantic top-down maps~\cite{chaplot2020object, zhang2021hierarchica, georgakis2021learning} to assist in waypoint planning. 
These methods, however, are limited to the specific object categories they were trained on and often depend on simulated data, hindering real-world application.

To overcome these limitations, zero-shot ObjectNav methods adopt frontier-based exploration.
Classical methods select exploration frontiers~\cite{yamauchi1997frontier} based on potential information gain~\cite{li2023learning}. 
Recent methods like CLIP on Wheels (CoW)~\cite{gadre2023cows} use CLIP features or an open-vocabulary object detector to identify target objects during frontier exploration.
Other works, such as LGX~\cite{dorbala2023can}, $\text{CM}^2$~\cite{georgakis2022cross}, ESC~\cite{zhou2023esc}, and StructNav~\cite{dragon}, incorporate large language models (LLMs) to score frontiers based on object detections. However, these approaches require converting visual features to text, introducing computational bottlenecks and often requiring remote server access.
In contrast, our approach leverages a language-driven semantic segmentation encoder, LSeg~\cite{lseg} that operates directly on RGB observations to generate semantic embeddings in the CLIP~\cite{clip} feature space.
This method eliminates the need for text generation from visual data and can run efficiently on consumer hardware, facilitating real-time zero-shot navigation.
Recent advancements include VLFM~\cite{vlfm}, which constructs occupancy maps from depth observations and uses a vision-language model to select exploration frontiers based on semantic values extracted from RGB images. This method facilitates open-vocabulary navigation, unlike earlier approaches that are restricted to a fixed set of object categories.
Our approach is different from these methods in integrating the imaginative capability of humans into robotic exploration. The Imagination Module predicts unseen parts of the environment using current and past observations, enhanced with semantic understanding via \textit{GeoSem Maps} - a vision-language grounded scene representation that consolidates geometry and semantics, providing a more robust and adaptable solution for object goal navigation in diverse environments.

\subsection{Scene Representation}
Effective scene representation is crucial for robotic navigation. Various methods have been proposed, such as ~\cite{garg2024robohop}, which uses segments from images as nodes in a topological map. ~\cite{jatavallabhula2023conceptfusion} and ~\cite{gu2023conceptgraphs} employ multi-modal scene representations combining CLIP~\cite{clip} and DINO~\cite{dino} embeddings to enhance 3D mapping and semantic understanding. These methods, however, require pre-exploration and mapping of the environment, limiting their applicability in real-time navigation tasks.
In \cite{navigating_real_world}, it is shown that modular approaches that use an intermediate scene representation outperform end-to-end approaches and transfer the best from sim to real, which further reinforces our decision to use \textit{GeoSem Maps} as an intermediate representation.
Apart from the visual language features, spatial information such as floor plans are often used as well in the navigation task. 
One method extends maps by 2 meters in all directions using generative models to address environmental uncertainty~\cite{katyal2019uncertainty}, but it falls short of producing complete floor plans, limiting scalability. 
Another approach predicts top-down semantic maps trained with randomly masked inputs~\cite{zhang2024imagine}, yet such masking does not mirror real-world observations. We instead simulate agents navigating environments to generate realistic partial views.
A different method infers unseen walls as 2D line segments from 360° LIDAR data~\cite{ericson2024beyond}, but treats floor plan prediction as a vision task with metrics misaligned with actual navigation.~\cite{pop3d} also treats the prediction of Open-Vocabulary 3D Occupancy BEV Maps in outdoor scenes as a computer vision task.
Our method evaluates imagination by simulating navigation performance of the agent in novel environments, offering a more practical and task-driven benchmark.

%% file: sec/3_Approach.tex
\section{Methodology}
\subsection*{Problem Formulation}
\label{sec:problem_formulation}
In this work, we address the Object Goal Navigation (ObjectNav) task~\cite{batra2020objectnav}, where an agent is tasked with locating an instance of a specified object category (e.g., `bed', `chair') within a previously unseen environment.
The agent is initialized at a random location and is provided with the target object category as an input goal.

At each time step \( t \), the agent receives visual observations, consisting of first-person RGB and depth images and pose readings, which indicate its position and orientation relative to its starting point.
The action space \( \mathcal{A} \) includes: \texttt{MOVE FORWARD} (0.25m), \texttt{TURN LEFT} (30°), \texttt{TURN RIGHT} (30°), \texttt{LOOK UP} (30°), \texttt{LOOK DOWN} (30°), and \texttt{STOP}.
The objective is for the agent to navigate towards the goal object and take the \texttt{STOP} action when it believes it has reached within 1 meter of the target, with a maximum of 500 steps allowed.

\subsection*{System Overview}
\begin{figure*}[htt]
    \centering
    \includegraphics[width=0.9\textwidth]{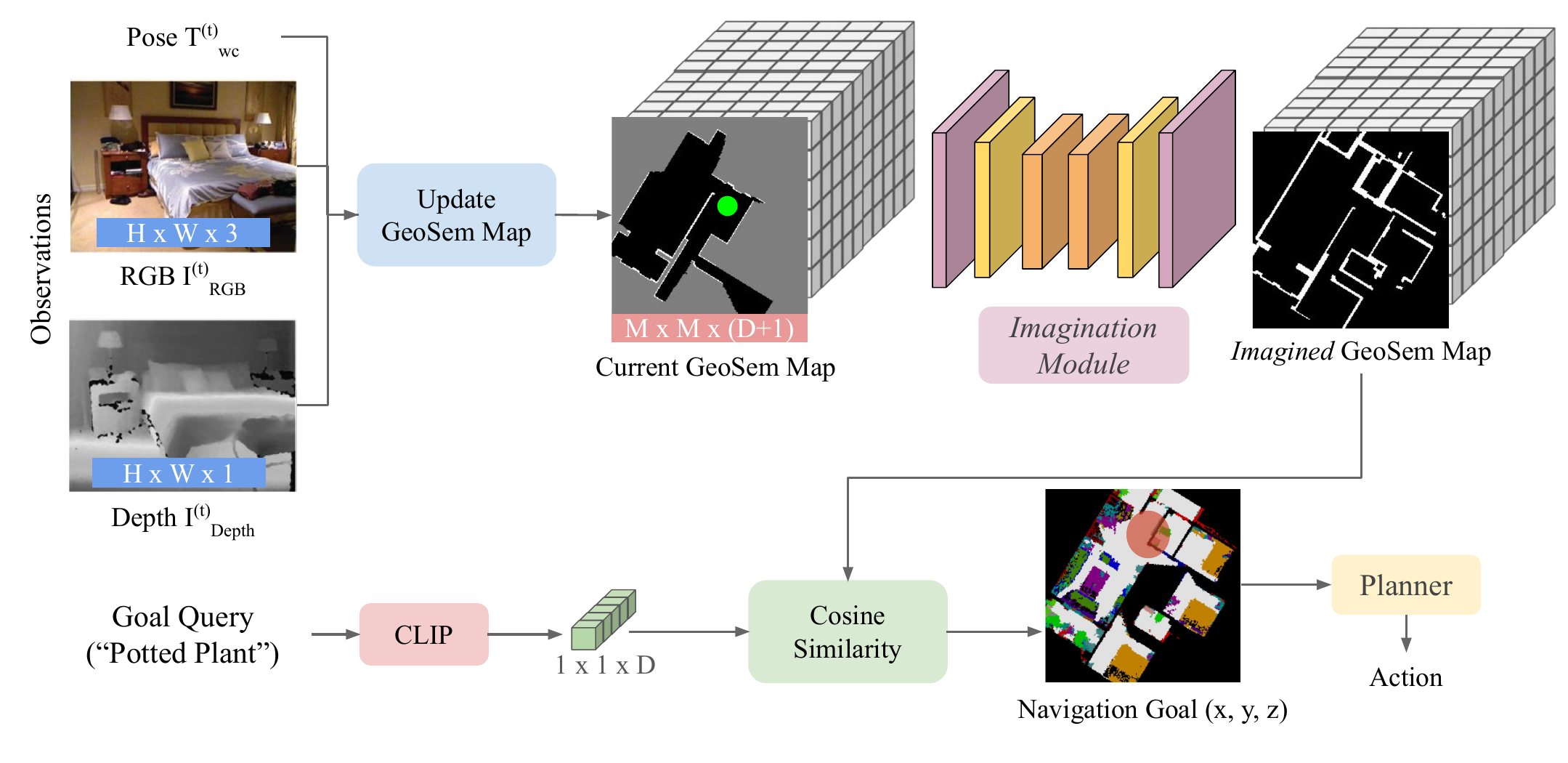}
    \caption{Overview of ForesightNav. At each timestep, we process the RGBD+Pose observations to update our current GeoSem Map, which acts as a database storing scene geometry and semantics. We use the \textit{Imagination Module} to predict the complete GeoSem Map of the scene and use the similarity scores between the Imagined GeoSem Map and the CLIP encoding of the goal query to compute the navigation goal. The planner then deterministically outputs actions to navigate to the goal. We explore only until the agent does not detect the target object.}
    \label{fig:overview}
\end{figure*}

Our method, shown in Figure \ref{fig:overview}, mainly consists of two building blocks: (i) \textbf{\textit{GeoSem Map}}: a scene representation that acts as a memory for the agent for jointly saving geometry and semantics from past observations.
In this work, we use Occupancy and CLIP~\cite{clip} as the geometry and semantic representations, respectively.  (ii) 
 \textbf{\textit{Imagination module}}: the exploration strategy for deducing the long-term navigation goal. As the agent moves through the environment, it records its observations in a GeoSem Map. The imagination module then predicts the complete GeoSem Map of the entire scene based on the current, partially observed one. Using this predicted GeoSem Map, the agent selects a long-term navigation goal to explore the path to the target object category efficiently. The agent subsequently uses the predicted geometry of the scene to navigate to the chosen goal. Once the target object category is within the field-of-view of the agent, it then directly navigates to the object without further exploration.

\subsection{GeoSem Map Construction}
\label{sec:geosem_construction}
A GeoSem Map is a  Bird’s-Eye-View (BEV) projection of 3D language-aligned voxel features.
It acts as a database for the agent to store inferred RGBD observations and to efficiently retrieve the information subsequently during the exploration, similar as done in~\cite{huang2023visual}.
Let \( \mathcal{X} \subset \mathbb{R}^3 \) represent the 3D spatial domain of the environment. 
Specifically, we define the GeoSem Map as a structured grid-based scene representation $
    \mathcal{M} \in \mathbb{R}^{H \times W \times (D+1)}$.
Here \( H \times W \) defines the spatial resolution of the grid, discretizing the environment into uniform cells.
\( D \) is the dimensionality of the CLIP embedding space.
The first \( D \) channels at each grid cell store \textbf{semantic features}, represented by a function:
    \begin{equation}
        f_{\text{sem}}: \mathcal{X} \to \mathbb{R}^D,
    \end{equation}
    which maps each spatial location to a $D$-dimensional CLIP feature vector.
The last channel stores \textbf{occupancy features}, defined as
    \begin{equation}
        f_{\text{occ}}: \mathcal{X} \to \{0,0.5,1\},
    \end{equation}
where \( f_{\text{occ}}(x) \in \{0, 0.5, 1\} \), representing free, unknown and occupied cells respectively. 
Thus, for a 3D point $p$ mapped to the grid cell \( (i,j) \), its representation is
\begin{equation}
    \mathcal{M}_{i,j} = \big[ f_{\text{sem}}(p) , f_{\text{occ}}(p) \big] \in \mathbb{R}^{D+1},
\end{equation}
where the first \( D \) elements of \( \mathcal{M}_{i,j} \) encode high-level semantic attributes, and the last element encodes occupancy. We initialize the first $D$ channels of $\mathcal{M}$ with zeros, and the $(D+1)^\text{th}$ channel with $0.5$ at the start of each episode, and update it as the episode progresses. 
\\

\textbf{Semantic Mapping }$\mathbf{f_{\text{sem}}}$: At each timestep \( t \), we compute a dense per-pixel semantic embedding using the LSeg~\cite{lseg} encoder
\begin{equation}
    f_{\text{LSeg}}: I_{\text{RGB}}^{(t)} \to \mathbb{R}^{H_I \times W_I \times D},
\end{equation}
where \( I_{\text{RGB}}^{(t)} \) is the input image, \( (H_I, W_I) \) are its dimensions, and each pixel is mapped to a \( D \)-dimensional CLIP-aligned embedding. Using the depth image \( I_{\text{Depth}}^{(t)} \), we project each pixel \( (u, v) \) to a local 3D point cloud
\begin{equation}
    p^{(t)}(u, v) = d(u,v) K^{-1} \begin{bmatrix} u \\ v \\ 1 \end{bmatrix},
\end{equation}
where \( d(u,v) \) is the depth value, and \( K \) is the intrinsic camera matrix. Each 3D point inherits the CLIP embedding of its corresponding pixel. To align with our global BEV representation, each point is transformed into the world frame using the camera pose \( T_{wc}^{(t)} \),
\begin{equation}
    p_w = T_{wc}^{(t)} p^{(t)}.
\end{equation}
These transformed points are then discretized into the grid map \( \mathcal{M} \) with spatial dimensions \( H \times W \). If multiple points fall into the same grid cell \( \mathcal{M}_{i,j} \), their embeddings are averaged, following the construction of ~\cite{huang2023visual}.
\begin{equation}
    \mathcal{M}_{i,j} = \frac{1}{N_{i,j}} \sum_{k \in \mathcal{S}_{i,j}} f_{\text{sem}}(p_k),
\end{equation}
where \( \mathcal{S}_{i,j} \) is the set of points mapped to grid cell \( (i,j) \), and $N$ forms the density map where \( N_{i,j} =  |\mathcal{S}_{i,j}|\).
\\

\textbf{Occupancy Mapping }$\mathbf{f_{\text{occ}}}$: 
To compute occupancy, we use the density map \( N \). To ensure relevance to the agent’s navigability, we apply a height-based filter based on the \textit{z}-coordinate of each point. Only points within a valid height range \( [z_{\min}, z_{\max}] \), determined by the agent type, are considered for occupancy calculation. A cell is then classified as occupied if its density exceeds 10\% of the maximum density across all grid cells.

\begin{figure}[b]
    \centering
    \begin{tabular}{ccc}
        \includegraphics[width=0.25\textwidth, trim={0 1cm 0 0},clip]{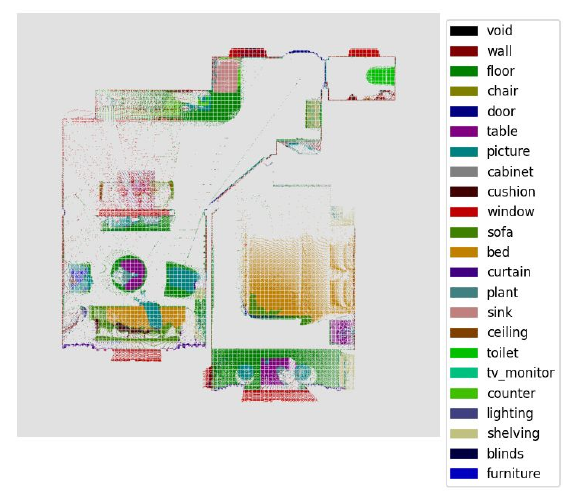} &
        \includegraphics[width=0.18\textwidth]{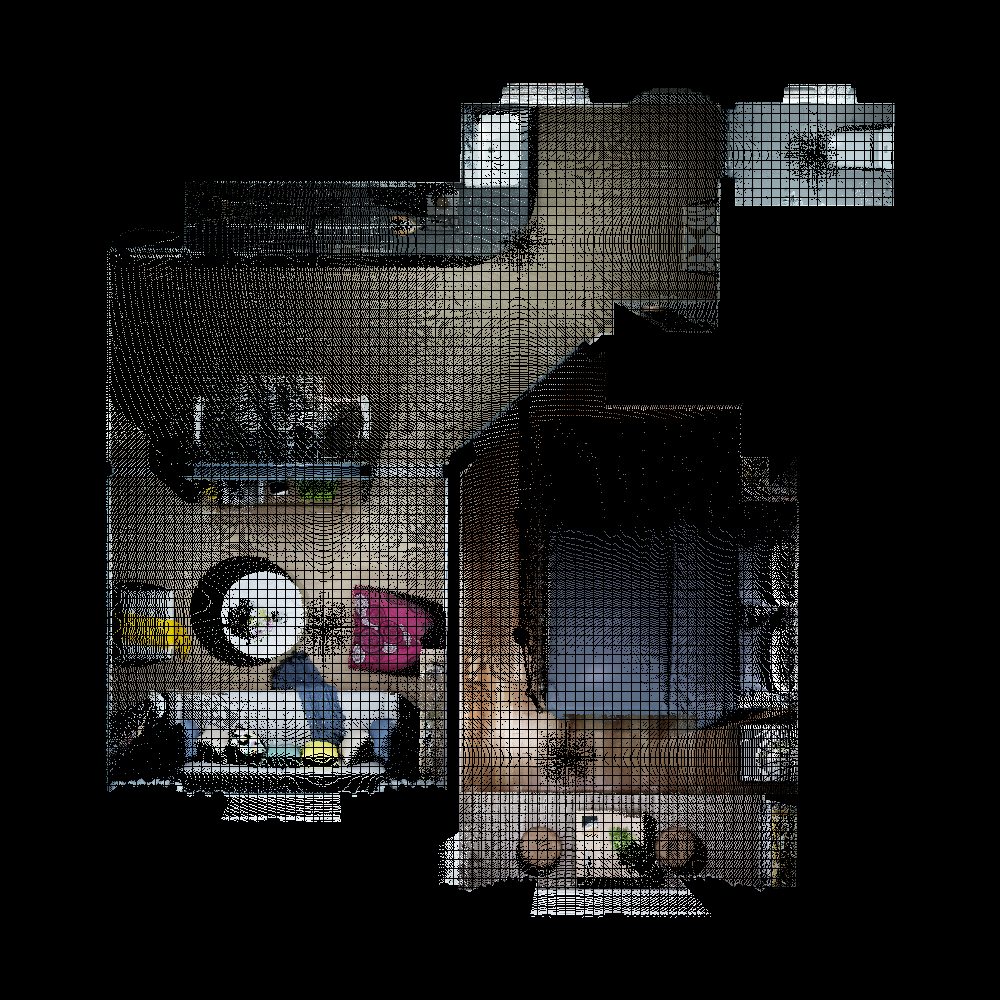} \\
        (a) Top-down semantic map & (b) Top-down color map
    \end{tabular}
    \caption{Visualization of a Structured3D scene: (a) Top-down semantic map with Matterport categories from the GeoSem Map, and (b) Top-down color map representing visual information.}
    \label{fig:clip_maps}
\end{figure}

\textbf{Zero-Shot Object
Segmentation: }In Figure \ref{fig:clip_maps} we show the top-down semantic map and the top-down color map of a scene from the Structured-3D dataset. To generate the top-down semantic map, given $K$ queries, we first encode each query using the CLIP text encoder, resulting in $K$ query vectors. We then compute the dot product of these query vectors with every cell$(H \times W)$ of the GeoSem Map $\mathcal{M}$, yielding $K$ similarity scores for each grid cell. The grid cell is then assigned to the category corresponding to the query with the highest similarity score.

\subsection{The Imagination Module}
\label{sec:imagination}
We define the Imagination Module as a learnable function that predicts the ground-truth GeoSem Map of a scene given a partially observed GeoSem Map. This involves predicting the complete occupancy and semantics of the scene. Formally, we define the \emph{Imagination Module} $f_{\text{imagine}}$ as
\begin{equation}
    f_{\text{imagine}}: \mathbb{R}^{H \times W \times (D+1)} \to \mathbb{R}^{H \times W \times (D+2)}.
\end{equation}
Given $\mathcal{J} = f_{\text{imagine}}(\mathcal{M})$, the input $\mathcal{M}$ is a GeoSem Map constructed using a partially observed scene where, 
\begin{equation}
\label{eq:input_clipmap}
    \mathcal{M}_{i,j} = \big[ o_\text{semantic} , o_{\text{occ}} \big] \in \mathbb{R}^{D+1}.
\end{equation}
Here $\mathbf{o_\text{semantic}},$ is the first \( D \) elements of \( \mathcal{M}_{i,j} \) encode already observed CLIP features of the scene. If $(i,j)$ is unobserved, $\mathcal{M}_{ij}[:D] = 0$.
And $\mathbf{o_{\text{occ}}} \in \{0,0.5,1\}$. 

The output $\mathcal{J}$ is again a 3D grid representation:
\begin{equation}
\label{eq:pred_clipmap}
    \mathcal{J}_{i,j} = \big[ p_\text{semantic} , p_{\text{occ}}, p_\text{interior} \big] \in \mathbb{R}^{D+2},
\end{equation}
where $\mathbf{p_\text{semantic}}$ is the first \( D \) elements of \( \mathcal{J}_{i,j} \) encode predicted high-level semantic attributes of the scene in the CLIP feature space (including the observed and unobserved cells).
$\mathbf{p_{\text{occ}}} \in [0,1],$ represents the $(D+1)^{\text{th}}$ element encodes the occupancy probability,
and $\mathbf{p_\text{interior}} \in [0,1],$ represents the $(D+2)^{\text{th}}$ element encodes the probability of a pixel lying inside the scene.
The need for $p_\text{interior}$ is justified later when we define the loss functions.
\\

\textbf{Generating Training Data: Partially Observed Occupancy Masks}
A training pair for the $f_\text{imagine}$ network consists of $(\mathcal{M}, \mathcal{G})$ where $\mathcal{M} \in \mathbb{R}^{H \times W \times (D+1)}$ is the observed GeoSem Map as defined in equation \ref{eq:input_clipmap} and $\mathcal{G} \in \mathbb{R}^{H \times W \times (D+2)}$ is the ground truth GeoSem-Map along with the ground truth interior mask of the scene, as defined in equation \ref{eq:pred_clipmap} with the true values instead of predicted scores. We can generate multiple instances of $\mathcal{M}$ from $\mathcal{G}$ by masking parts of the occupancy channel $\mathcal{G}_{m_i,m_j,D+1} = 0.5$ and the semantic channels $\mathcal{G}_{m_i,m_j,1:D} = 0$, where $(m_i,m_j) \in \text{O}$ and $O$ is a mask. To ensure that we train $f_\text{imagine}$ on instances that it will encounter during inference for the task of navigation of an agent in an indoor environment, we generate multiple masks $O$ for every scene in the training split of the dataset by simulating an agent in the scene.

\begin{figure}[ht]
\centering
\includegraphics[width=0.475\textwidth]{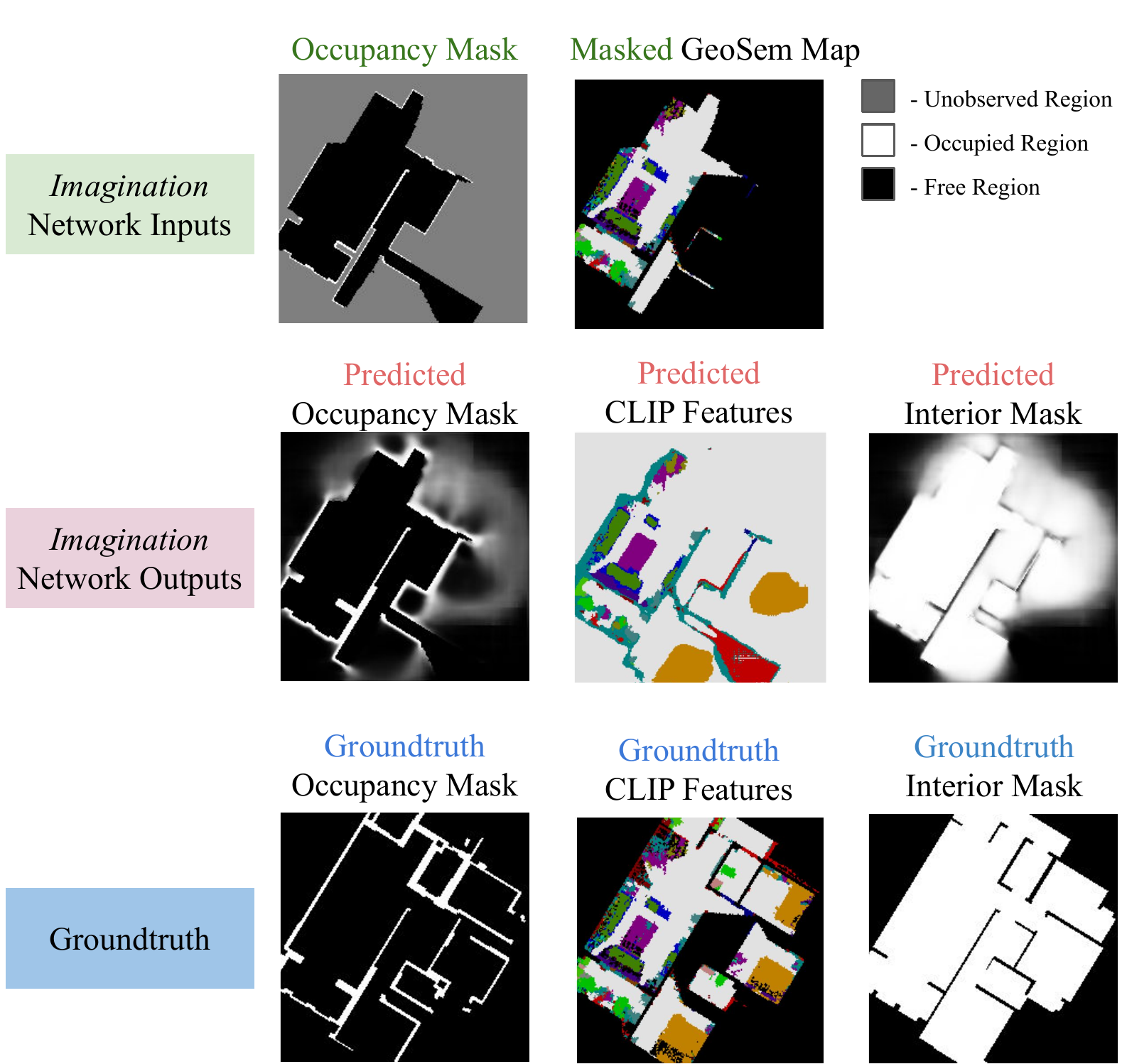}
\caption{Example of inputs, predictions, and the associated ground truths for the Imagination Module. The GeoSem Map is a multi-dimensional tensor, but for visualization purposes, we show the top-down semantic map generated using the Matterport object categories as queries.}
\label{fig:input_output_maps}
\end{figure}

For each training scene, we start by simulating an agent in a 2D grid world - The ground truth occupancy map of each scene, denoted $\mathbf{C}$. The agent is defined with specific parameters such as `field of view (FOV)', `sensor range', and `turn angle', which mimic real-world sensor and agent configurations. The agent is randomly initialized at an interior point of the scene and is tasked with reaching randomly generated waypoints that are distributed all over the scene using the A* algorithm. The agent maintains a `predicted map' $\mathbf{P}$ initialized with all cells marked as unobserved ($0.5$). At each timestep $t$, the agent simulates a depth sensor (LIDAR) to gather occupancy information within its specified FOV and orientation. The agent collects depth data up to the maximum 'sensor range' distance, updating $\mathbf{P}_t$ accordingly. The observations at each timestep are derived from the ground truth occupancy map $\mathbf{C}$,
\begin{equation}
\mathbf{P}_{t+1} = \mathbf{P}_t \odot (1 - \mathbf{O}_t) + \mathbf{C} \odot \mathbf{O}_t,
\end{equation}
where $\mathbf{O}_t$ is the observation mask at timestep $t$ indicating the regions observed by the agent. 
Figure \ref{fig:input_output_maps} illustrates an overview of the inputs, outputs, and ground truths for the Imagination Module.
\\

\textbf{Network Architecture and Training Setup: }To make the training process robust to various real-world scenarios, we apply on-the-fly data augmentations such as rotation and scaling to the training samples. We explore two different architectures for the imagination network: a Vision Transformer (ViT)~\cite{vit} based architecture using Masked Autoencoders (MAE) and a Convolutional Neural Network (CNN) based architecture using a U-Net~\cite{unet}.
We supervise the imagination network using three loss components: CLIP embedding similarity, occupancy prediction, and interior mask prediction.

For \textit{CLIP Embedding Similarity}, we use the cosine similarity loss for the CLIP embeddings, applied only to the interior regions since it will be undefined for exterior regions with the CLIP-Embeddings set to zero vectors,
\[
\mathcal{L}_{\text{CLIP}} = \frac{1}{|\Omega_{\text{int}}|} \sum_{(i,j) \in \Omega_{\text{int}}} \left( 1 - \frac{\mathcal{G}_{i,j,1:D} \cdot \mathcal{J}_{i,j,1:D}}{\|\mathcal{G}_{i,j,1:D}\| \|\mathcal{J}_{i,j,1:D}\|} \right),
\]
where $\Omega_{\text{int}}$ is the set of interior grid cells, $\mathcal{J}$ is the $f_{\text{imagine}}$ output and $\mathcal{G}$ is the groundtruth. Since the exterior regions of the scene do not receive any supervision, the network tends to output random CLIP-embedding values for these regions. To ensure that these do not affect the long-term navigation goal selection, we also predict the interior mask and only look for the navigation goal within the predicted interior mask.

The occupancy prediction loss 
is defined as
\begin{equation}
\mathcal{L}_{\text{occ}} = \text{wBCE}(\mathcal{J}_{i,j,D+1} , \mathcal{G}_{i,j,D+1}),
\end{equation}
which is a weighted binary cross-entropy (BCE) loss to account for the sparsity of occupied cells as compared to unoccupied cells. We also ablate by treating occupancy prediction as a regression problem by using the MSE loss for supervision. The interior mask is supervised using a standard BCE loss $\mathcal{L}_{\text{interior}} = \text{BCE}(\mathcal{J}_{i,j,D+2} , \mathcal{G}_{i,j,D+2})$.
Thus, the final formulation of the multi-task loss is as
\begin{equation}
    \mathcal{L} = \lambda_{\text{CLIP}}\cdot\mathcal{L}_{\text{CLIP}} + \lambda_{\text{occ}}\cdot\mathcal{L}_{\text{occ}} + \lambda_{\text{interior}}\cdot\mathcal{L}_{\text{interior}}.
\end{equation}

\subsection{Long-Term Navigation Goal Extraction}

\begin{figure}[t]
    \centering
    \includegraphics[width=0.48\textwidth]{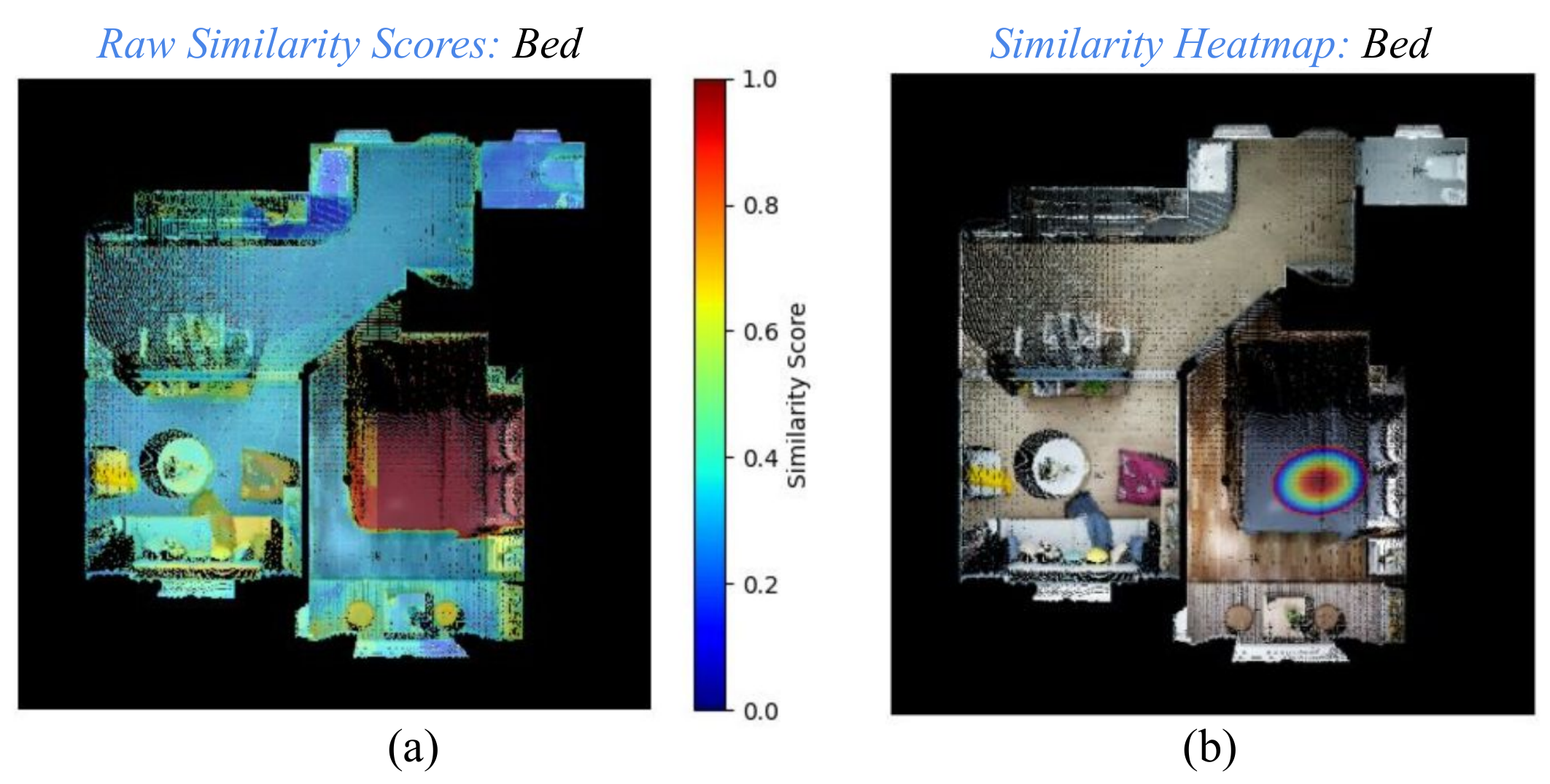}
    \caption{Visualization of query spatial distribution: (a) Raw similarity scores heatmap overlaid on a top-down RGB map showing potential locations of the query. (b) Gaussian Mixture Model (GMM) covariances as ellipses and log-likelihood heatmap overlaid on the top-down color map, visualizing the spatial distribution of the query.}
    \label{fig:query_distribution}
    \vspace{-1em}
\end{figure}

\begin{figure}[t]
    \centering
    \includegraphics[width=0.48\textwidth]{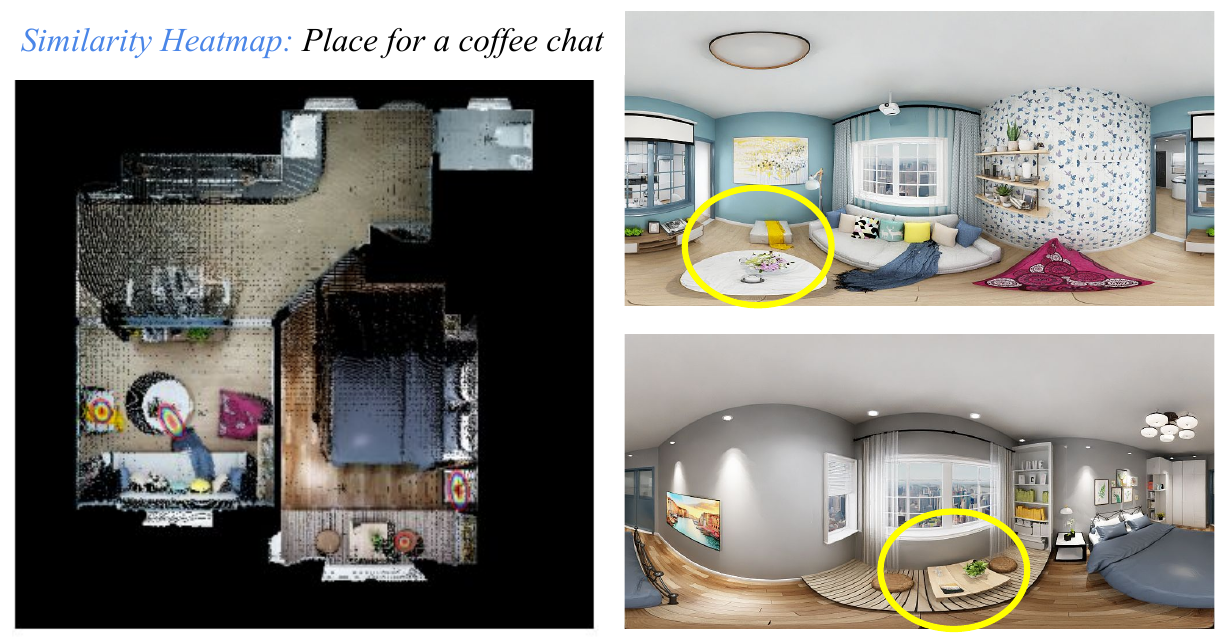}
    \caption{An example of querying an open-vocabulary language query in a GeoSem Map: "\textit{Place for a Coffee Chat}". On the left, we show the heatmap with GMM covariances and on the right the true locations of the query in the scene.}
    \label{fig:openvocab_performance}
    \vspace{-1em}
\end{figure}

The task of exploration during ObjectNav can be summarized as choosing the point $(x,y,z)$ to navigate to, given the past observations, so that the target object can be found.
In our approach, we use the predicted GeoSem-Map output from $f_{\text{imagine}}$ to decide the navigation goal. Since we make use of CLIP features, we can use any language query to navigate to, and the query does not need to be from a specified set of target categories. Given $\mathcal{J} = f_{\text{imagine}}(\mathcal{M}_t)$, where $\mathcal{M}_t$ is the GeoSem Map constructed from observations from $(0,t]$, we choose the navigation goal using the cosine similarity heatmap between $\mathcal{J}[\mathcal{J}_{:,:,D+2}>0.5]$ and $\text{CLIP}_{\text{text}}(\text{Goal\_Query})$ - only for the pixels that lie inside the interior mask. Ideally, the pixel in the heatmap with the highest score should be the goal. To minimize false positives and ensure the robustness of the goal location, and to account for multiple occurrences of the query, we process the heatmap further. We first normalize and threshold the similarity map to get a binary heatmap indicating potential goal locations. Next, we apply some outlier removal using DBSCAN and find the optimal number of clusters ($K$) using the silhouette score metric. We then fit a Gaussian Mixture Model (GMM) with $K$ gaussians to get the final candidate zones for the goal query, visualized in Figure \ref{fig:query_distribution}. In Figure \ref{fig:openvocab_performance}, the performance on an open-vocabulary language query is visualized. We convert the coordinates of the location from grid to world for the planner, and we choose the closest cluster to the current robot position when $K>1$. 

Imagination and goal selection are triggered at the same frequency - whenever a new \textit{GeoSem Map} is predicted, a new navigation goal is chosen. The map is updated when the agent's current field of view has low overlap with the field of view at the last update, based on a predefined threshold.


%% file: sec/4_Experiments.tex
\section{Experiments and Results}
\label{sec:experiments}

\textbf{Dataset: }For the training of the Imagination module and evaluation of our method, we use scenes from the Structured3D dataset~\cite{zheng2020structured3d}, which is a large-scale photo-realistic dataset with 3500 scenes. We train the Imagination module on the train-val split recommended by the dataset authors i.e. 3000 / 500. Each room in a scene has panoramic RGB and depth images, semantic annotations, and ground truth junction information, which makes it easy to construct a floorplan of the room from the annotations, providing the ground truth occupancy map for walls. This makes Structured3D an ideal candidate for training the imagination layer of our exploration strategy. With this work we introduce the Structured3D benchmark for ObjectNav. All evaluations are reported on the validation split of Structured3D, which we do not use for training the Imagination Module.

\textbf{Network Details: }We use $H=224, W=224$ as the spatial dimensions for the GeoSem Map, with $z_{\text{min}}=30\%$ and $z_{\text{max}}=70\%$ of the scene height to emulate an occupancy map for a drone agent. The \textit{ViT-B/32} model variant of CLIP is used, which gives us $D=512$ dimensional embeddings, making the size of our GeoSem Map $224 \times 224 \times 513$. The convolution depths for the UNet architecture are $[64, 128, 256, 512, 1024]$ and the threshold to convert the predicted occupancy probability to a binary occupancy map for navigation is set to $0.72$. We also ablate on using the \textit{mae\_vit\_base\_patch16}~\cite{he2022masked} ViT model for the Imagination module. The loss coefficients were set as $\lambda_{\text{CLIP}}=10,\lambda_{\text{occ}}=1, \lambda_{\text{interior}=1}$.

\textbf{Evaluation Setup: }The Structured3D is set up for the closed-loop evaluation task of ObjectNav as defined in Section \ref{sec:problem_formulation}. To evaluate the performance of ObjectNav, we employ three key metrics as proposed by~\cite{objnav_eval}: Success, Success weighted by Path Length (SPL), and Distance to Goal. An agent is 
randomly initialized in a grid world, and its observations are simulated using the groundtruth GeoSem Map of the scene, which has been precomputed. The simulation for this evaluation occurs in a similar manner to the generation of occupancy masks in Section \ref{sec:imagination}, where an agent navigates a 2D grid world using predefined parameters such as field of view (fov), sensor range, and turn angle.

\subsection{Evaluating on Occupancy Prediction}
Before we evaluate our approach on ObjectNav, we experiment with the Imagination strategy on the task of PointNav (Point Goal Navigation), where the task is to reach a point-goal $(x,y,z)$ in an unseen environment. The agent starts in a random position and it is given two goal waypoints $(x_1, y_1)$ and $(x_2, y_2)$ to reach in order, with the simulation running for a maximum of 3000 timesteps. We compare the performance of two agents, both using the A* algorithm for planning:\\
\begin{enumerate}
    \item \textit{Imagination Agent}: Every 10 timesteps, the imagination agent imagines the occupancy map using the imagination layer and the A* planner uses this imagined map to plan its path to the goal.
    \item \textit{Vanilla Agent}: The vanilla agent replans using the current occupancy map every 10 timesteps.
\end{enumerate}

In Table~\ref{table:closed_loop_simulation}, we report each method's completion rate and mean timesteps. We based our analysis on the network architecture and the loss we used to supervise the Imagination Module by treating occupancy prediction as a regression(MSE Loss) and classification(BCE Loss) task.

\begin{table}[ht]
\centering
\fontsize{9pt}{12pt}\selectfont
\begin{tabular}{lcc}
\toprule
\makecell{\textbf{Method}} & \makecell{\textbf{Completion Rate} ($\uparrow$)} & \makecell{\textbf{Mean} ($\downarrow$)} \\ \hline
Vanilla & 0.99 & 640.65 \\
\rowcolor{green!12}UNet + BCE & \textbf{1.0} & \textbf{439.38} \\
UNet + MSE & \textbf{1.0} & 448.23 \\
ViT + BCE & 0.85 & 465.72 \\
ViT + MSE & 0.89 & 499.78 \\ 
\bottomrule
\end{tabular}
\caption{Closed-loop simulation evaluation results. Arrows indicate whether higher ($\uparrow$) or lower ($\downarrow$) values are better. Bold values indicate the best performance for each metric.}
\label{table:closed_loop_simulation}
\vspace{-1em}
\end{table}

\begin{figure}[ht]
\centering
\includegraphics[width=0.5\textwidth]{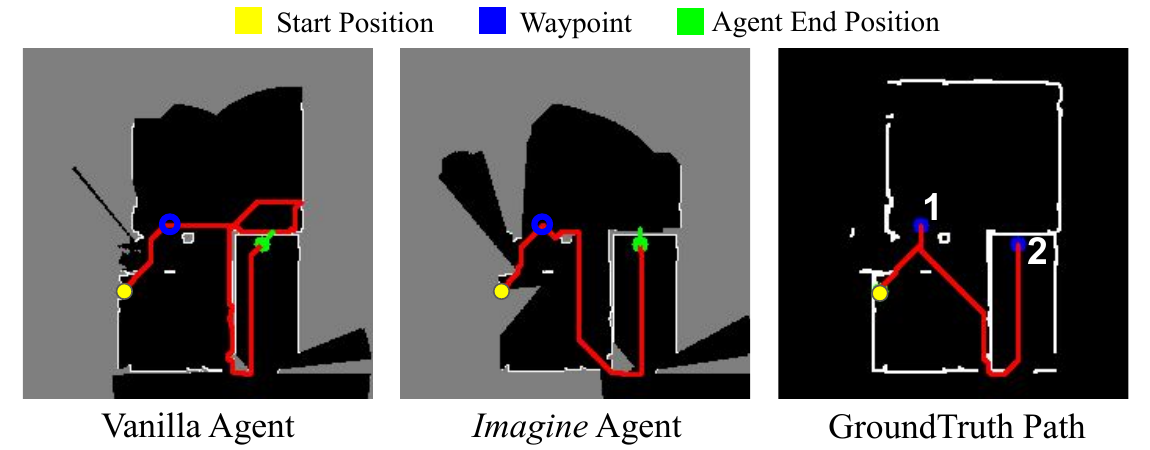}
\caption{Paths taken by the vanilla agent and the imagine agent during a run. The imagine agent predicts the structure of the scene and selects a more efficient path. In the first two images, the green circle denotes the final position of the agents after reaching the goals. The last image (groundtruth path) shows the starting position of the agent in green, and the two goal waypoints in blue.}
\label{fig:paths_comparison}
\end{figure}

The imagination agents outperform the vanilla agent with the UNet Imagination backbone supervised using BCE Loss, which gives the best performance, providing close to a \textbf{1.5x speedup} in reaching the goals compared to the vanilla agent. This result shows that predicting the walls from partially observed occupancy maps helps the agent navigate faster. Figure \ref{fig:paths_comparison} visualizes the paths taken by the vanilla agent and the imagine agent for a particular run. After reaching the first waypoint, the imagine agent accurately predicts the structure of the scene and chooses the path from the south, which is closer to the ground truth best path, unlike the vanilla agent that attempts to reach the goal from the north.

This observation backs our hypothesis that employing an approach backed by human reasoning for exploration can yield substantial gains in navigation efficiency. The ability of the imagine agent to anticipate the layout of unobserved regions and adapt its path accordingly demonstrates the potential of integrating imagination layers in navigation tasks. \\

\subsection{Evaluating Imagination for ObjectNav}

We evaluate our Imagination exploration strategy against other recent baselines. We choose frontier-based exploration techniques as baselines for a fair evaluation since our method is modular-learning based. Defined in~\cite{navigating_real_world}, modular-learning approaches use an intermediate scene representation to compute the navigation goal as opposed to end-to-end approaches that directly make predictions in the action space. We evaluate and report metrics for each method for each category of the ObjectNav task as described in~\cite{objnav_eval}, namely: [``chair", ``couch", ``potted plant", ``bed",``toilet", ``tv"] with an episode timeout of $500$s. For all agents, when the goal coordinate appears in their observation, i.e., $P[g_x, g_y] \neq 0.5$ where $P$ is the current occupancy map, it is assumed that the agent can detect the goal category and the exploration phase is over and the agent deterministically navigates to the goal.

\textit{Baselines: }All baselines are frontier-based, and hence, at each timestep, the agent needs to decide which frontier to choose for exploration, which becomes the navigation goal for the planner.
\begin{enumerate}
    \item \textit{VLFM-CLIP}: VLFM~\cite{vlfm} agent using CLIP for language grounding of features instead of BLIP-2~\cite{li2023blip}. 

    \item \textit{StructNav-Frontiers}: Inspired by the approach of StructNav~\cite{dragon}, this baseline uses a top-down semantic map with $K$ fixed object categories (as introduced in~\cite{chaplot2020object}) as the scene representation instead of scene graphs proposed in the original method. The agent uses a prior distance matrix - a $K \times K$ matrix that stores similarity scores between the CLIP-Embedding vectors of the $K$ classes. The frontier with the highest score indexed from the prior matrix using the semantic category at the frontier and the goal category, is chosen as the navigation goal.

    \item \textit{Random Agent}: Chooses a random frontier as the navigation goal in each timestep.

    \item \textit{Greedy Agent}: Chooses the frontier that is closest to its current position at the current timestep.
\end{enumerate}

\begin{table}[ht]
\centering
\fontsize{9pt}{12pt}\selectfont
\begin{tabular}{lccc}
\toprule
\textbf{Method} & \textbf{SPL} ($\uparrow$) & \makecell{\textbf{Distance} \\ \textbf{To Goal} ($\downarrow$)} & \textbf{Success} ($\uparrow$) \\ \midrule
\rowcolor{green!12} ForesightNav (\textbf{Ours}) & \textbf{0.67} & \textbf{25.32} & \textbf{0.73} \\
VLFM-CLIP & 0.66 & 27.36 & 0.71 \\
StructNav-Frontiers & 0.63 & 33.38 & 0.66 \\
Random Agent & 0.62 & 29.30 & 0.68 \\
Greedy Agent & 0.58 & 37.06 & 0.60 \\ \bottomrule
\end{tabular}
\caption{Average metrics for different exploration strategies in the ObjectNav task. Our proposed approach outperforms existing state-of-the-art approaches on the  Structure3D validation dataset in three different evaluation metrics.}
\label{table:objectnav_results}
\end{table}

We report ObjectNav metrics averaged across the validation scenes of the Structured3D dataset for each of the ObjectNav categories in Table \ref{table:objectnav_results}.

By leveraging imagination-driven exploration, \textbf{ForesightNav} achieves absolute increases in the success rates weighted by path length (SPL) over prior SOTA approaches. This improvement is driven by its ability to anticipate unseen regions, effectively guiding the agent towards high-reward areas without unnecessary exploration.
This suggests that imagination aided by GeoSem Maps enhance the agent's ability to infer promising frontiers, prioritizing those most likely to contain the target object. This capability is particularly valuable in large or complex environments where exhaustive search is computationally expensive.  

While StructNav-Frontiers also leverages semantic information, it is outperformed by ForesightNav and VLFM-CLIP due to the semantic richness that CLIP embeddings provide. The Random Agent and Greedy Agent predictably perform the worst, further emphasizing the critical role of advanced scene representation and the exploration strategy in efficient navigation.

%% file: sec/5_Conclusion.tex
\section{Conclusion}  
In this work, we have presented \textit{ForesightNav}, a novel exploration strategy for Object Goal Navigation that leverages the power of data-driven imagination. By integrating CLIP’s visual-semantic understanding with an anticipatory reasoning framework, our method enables an agent to infer unexplored regions and make informed navigation decisions, leading to superior performance over prior SOTA approaches. Our results demonstrate that \textit{ForesightNav} significantly improves navigation efficiency for both the PointNav and ObjectNav tasks, also achieving higher success rates. 

Our findings suggest that data-driven imagination is the next step in robotic exploration, with the potential to redefine how agents navigate complex environments. By training agents to \textit{imagine} spaces before seeing them, we move closer to creating foundation models for indoor scene understanding, similar to how large vision-language models have revolutionized perception.

\paragraph{Limitations and Future Work:} 
While the imagination strategy we propose is effective in indoor environments with a BEV-style GeoSem map, for navigating large scenes using non-object-centric language queries - a higher resolution map must be used, which will significantly increase memory usage and the network size of the imagination model. A hierarchical memory structure that builds on top of GeoSem Maps and merges it with a structure like 3D Scene Graphs~\cite{armeni20193d} for faster retrieval is a possible extension to this work.

%% file: sec/6_Acknowledgement.tex
\section{Acknowledgements}
This work was supported by the European Union’s Horizon Europe Research and Innovation Programme under grant agreement No. 101120732 (AUTOASSESS) and the European Research Council (ERC) under grant agreement No. 864042 (AGILEFLIGHT).

%% file: sec/X_suppl.tex
\setcounter{page}{1}
\maketitlesupplementary

\begin{multicols*}{2}


\section{GeoSem Maps from Equirectangular RGBD + Pose}
\label{sec:equirectangular_geosem}
The Structured3D dataset~\cite{zheng2020structured3d} provides richly annotated 3D indoor scenes with panoramic RGBD observations. It contains 3,500 scenes in total, split into 3,000 for training and 500 for validation. Each scene $S_i$ is composed of multiple rooms, and for every room $R_{ij}$, the dataset offers a set of equirectangular RGB images $I_{\text{equirect}_{ij}} \in \mathbb{R}^{H \times W \times 3}$, corresponding equirectangular dense depth maps $D_{\text{equirect}_{ij}} \in \mathbb{R}^{H \times W \times 1}$, and global camera poses $T_{ij} \in SE(3)$ for each panoramic viewpoint.

In this section, we describe the construction of \textit{GeoSem Maps} for each scene, which serve as groundtruth supervision for the imagination module discussed in Section~\ref{sec:imagination}. A representative Figure \ref{fig:perspective_images} depicts the overview of the method. Directly using the equirectangular images $I_{\text{equirect}}$ as input to the pre-trained LSeg encoder leads to degraded CLIP embeddings due to the geometric distortions introduced by the panoramic format. To mitigate this, we instead apply a perspective projection strategy, simulating standard pinhole camera views to generate high-quality, per-pixel CLIP embeddings.

\subsection{Perspective Projection for GeoSem Map Construction}
We utilize a perspective camera model with a field of view (FOV) of $90$ degrees. The viewing directions are strategically chosen so that the complete room is covered:
horizontal view directions $\Theta = [-180^\circ, -90^\circ, 0^\circ, 90^\circ]$ and vertical view directions $\Phi = [-45^\circ, 0^\circ, 45^\circ]$. Each combination of $\Theta$ and $\Phi$ results in a distinct perspective image capturing the room from different orientations i.e. $12$ viewing directions. Given an equirectangular image $I_{\text{equirect}_{ij}}$ of scene $S_i$ and room $R_{ij}$, perspective projection gives us $12$ images $I_{\text{persp}_{ijk}} \text{ where } k \in [1,12]$. With perspective projection, we loose the pixel-wise depth association between $I_{\text{equirect}}$ and $D_{\text{equirect}}$. To finally construct a GeoSem Map as described in \ref{sec:geosem_construction}, we need pixel-wise CLIP embeddings and depth, along with camera pose for top-down projection onto a BEV map.

\subsection{2D-3D Correspondence Using Indexing Approach}
Each pixel $(u, v)$ in $D_{\text{equirect}_{ij}}$ represents a depth value $d_{uv}$, which can be projected into a local 3D point $\mathbf{p}_{uv}^{\text{local}} \in \mathbb{R}^3$ using the intrinsic parameters of the equirectangular projection. These local 3D points are then transformed into the global coordinate frame using the camera pose:
\[
\mathbf{p}_{uv}^{\text{global}} = T_{ij} \cdot \mathbf{p}_{uv}^{\text{local}}
\]

A global point cloud is constructed for the scene by aggregating all such transformed 3D points across rooms:
\[
\mathcal{P}_i = \bigcup_{j} \{ \mathbf{p}_{uv}^{\text{global}} \mid (u,v) \in D_{\text{equirect}_{ij}} \}
\]

To retain the pixel-to-point correspondence for associating the CLIP embeddings later, an indexing scheme is used:

\begin{itemize}
    \item \textbf{Base Indexing:} Each panoramic image $I_{\text{equirect}_{ij}}$ is assigned a unique base index $B_{ij}$.
    
    \item \textbf{Pixel Indexing:} We create the index map $\texttt{\textbf{index}}_{ij} \in \mathbb{R}^{H \times W \times 1}$, where each pixel $(u,v)$ receives a unique global index within the scene:
    \[
    \texttt{\textbf{index}}_{ij}(u,v) = B_{ij} + (u \cdot W + v)
    \]
\end{itemize}
This ensures that every pixel across all panoramas in a scene has a globally unique identifier.

\paragraph{Perspective Projection and Index Transformation:} When we convert $I_{\text{equirect}_{ij}}$ into a perspective image $I_{\text{persp}_{ijk}}$, we similarly project the index map $\texttt{\textbf{index}}_{ij}$ to obtain a corresponding \textit{index perspective image} $\phi_{ijk}$ which allows us to define the mapping:

\[
\phi_{ijk}: \text{Pixels in } I_{\text{persp}_{ijk}} \rightarrow \text{Point IDs in } \mathcal{P}_i
\]
\[
\phi_{ijk}(u', v') = \texttt{\textbf{index}}_{ij}(u,v)
\]

This enables each pixel in the perspective image to retain the identity of its original 3D point, thereby preserving the connection between CLIP embeddings (obtained from LSeg) and their corresponding 3D positions in the scene.
\[
\text{LSeg}(I_{\text{persp}_{ijk}})(u', v') \to \mathcal{P}_i[\phi_{ijk}(u', v')]
\]

\subsection{CLIP Embedding Generation and Mapping}

With the perspective images and corresponding depth information associated using the index images, we employ the LSeg encoder to derive pixel-wise CLIP embeddings for each perspective image. 

The subsequent steps involve aggregating these CLIP embeddings into a grid representation $\mathcal{M}$, where each grid cell encapsulates semantic information from the underlying 3D points and their associated CLIP embeddings. The construction of the GeoSem Map $\mathcal{M}$ proceeds as outlined in Section \ref{sec:geosem_construction}.

\subsection{Handling Spurious Observations}

From the Structured3D annotations, the groundtruth floor plan is known. However, to better simulate real-world conditions during training, we generate the groundtruth occupancy maps using a top-down projection of the global pointcloud instead of using the dataset floor plan annotation. Due to imperfections in the point cloud, the resulting occupancy maps may contain small holes in the walls. These can cause "leaks" during agent simulation (described in Section~\ref{sec:experiments}) where depth rays may erroneously pass through walls. This leads to artifacts in the occupancy map - specifically, observed regions outside the intended scene boundaries which would not occur in a realistic setting. To address this, we generate an interior-exterior mask $\mathbf{E} \in \{0, 1\}^{M \times M}$ from the Structured3D annotations for room polygons, marking all points in the exterior as 0.5 (unobserved) in the occupancy map $P$:

\begin{equation}
    \mathbf{P} = \mathbf{P} \odot \mathbf{E} + 0.5 (1 - \mathbf{E})
\end{equation}

\end{multicols*}

%% file: main.bbl
\begin{thebibliography}{44}
\providecommand{\natexlab}[1]{#1}
\providecommand{\url}[1]{\texttt{#1}}
\expandafter\ifx\csname urlstyle\endcsname\relax
  \providecommand{\doi}[1]{doi: #1}\else
  \providecommand{\doi}{doi: \begingroup \urlstyle{rm}\Url}\fi

\bibitem[Anderson et~al.(2018)Anderson, Chang, Chaplot, Dosovitskiy, Gupta, Koltun, Kosecka, Malik, Mottaghi, Savva, et~al.]{objnav_eval}
Peter Anderson, Angel Chang, Devendra~Singh Chaplot, Alexey Dosovitskiy, Saurabh Gupta, Vladlen Koltun, Jana Kosecka, Jitendra Malik, Roozbeh Mottaghi, Manolis Savva, et~al.
\newblock On evaluation of embodied navigation agents.
\newblock \emph{arXiv preprint arXiv:1807.06757}, 2018.

\bibitem[Armeni et~al.(2019)Armeni, He, Gwak, Zamir, Fischer, Malik, and Savarese]{armeni20193d}
Iro Armeni, Zhi-Yang He, JunYoung Gwak, Amir~R Zamir, Martin Fischer, Jitendra Malik, and Silvio Savarese.
\newblock 3d scene graph: A structure for unified semantics, 3d space, and camera.
\newblock In \emph{Proceedings of the IEEE/CVF international conference on computer vision}, pages 5664--5673, 2019.

\bibitem[Batra et~al.(2020)Batra, Gokaslan, Kembhavi, Maksymets, Mottaghi, Savva, Toshev, and Wijmans]{batra2020objectnav}
Dhruv Batra, Aaron Gokaslan, Aniruddha Kembhavi, Oleksandr Maksymets, Roozbeh Mottaghi, Manolis Savva, Alexander Toshev, and Erik Wijmans.
\newblock Objectnav revisited: On evaluation of embodied agents navigating to objects.
\newblock \emph{arXiv preprint arXiv:2006.13171}, 2020.

\bibitem[Caron et~al.(2021)Caron, Touvron, Misra, J{\'e}gou, Mairal, Bojanowski, and Joulin]{dino}
Mathilde Caron, Hugo Touvron, Ishan Misra, Herv{\'e} J{\'e}gou, Julien Mairal, Piotr Bojanowski, and Armand Joulin.
\newblock Emerging properties in self-supervised vision transformers.
\newblock In \emph{Proceedings of the IEEE/CVF international conference on computer vision}, pages 9650--9660, 2021.

\bibitem[Chang et~al.(2020)Chang, Gupta, and Gupta]{chang2020semantic}
Matthew Chang, Arjun Gupta, and Saurabh Gupta.
\newblock Semantic visual navigation by watching youtube videos.
\newblock \emph{Advances in Neural Information Processing Systems}, 33:\penalty0 4283--4294, 2020.

\bibitem[Chaplot et~al.(2020)Chaplot, Gandhi, Gupta, and Salakhutdinov]{chaplot2020object}
Devendra~Singh Chaplot, Dhiraj~Prakashchand Gandhi, Abhinav Gupta, and Russ~R Salakhutdinov.
\newblock Object goal navigation using goal-oriented semantic exploration.
\newblock \emph{Advances in Neural Information Processing Systems}, 33:\penalty0 4247--4258, 2020.

\bibitem[Chen et~al.(2023)Chen, Li, Kumar, Ghanem, and Yu]{dragon}
Junting Chen, Guohao Li, Suryansh Kumar, Bernard Ghanem, and Fisher Yu.
\newblock How to not train your dragon: Training-free embodied object goal navigation with semantic frontiers.
\newblock \emph{arXiv preprint arXiv:2305.16925}, 2023.

\bibitem[Deitke et~al.(2022)Deitke, VanderBilt, Herrasti, Weihs, Ehsani, Salvador, Han, Kolve, Kembhavi, and Mottaghi]{procthor}
Matt Deitke, Eli VanderBilt, Alvaro Herrasti, Luca Weihs, Kiana Ehsani, Jordi Salvador, Winson Han, Eric Kolve, Aniruddha Kembhavi, and Roozbeh Mottaghi.
\newblock Procthor: Large-scale embodied ai using procedural generation.
\newblock \emph{Advances in Neural Information Processing Systems}, 35:\penalty0 5982--5994, 2022.

\bibitem[Dorbala et~al.(2023)Dorbala, Mullen~Jr, and Manocha]{dorbala2023can}
Vishnu~Sashank Dorbala, James~F Mullen~Jr, and Dinesh Manocha.
\newblock Can an embodied agent find your "cat-shaped mug"? llm-based zero-shot object navigation.
\newblock \emph{IEEE Robotics and Automation Letters}, 2023.

\bibitem[Dosovitskiy et~al.(2020)Dosovitskiy, Beyer, Kolesnikov, Weissenborn, Zhai, Unterthiner, Dehghani, Minderer, Heigold, Gelly, et~al.]{vit}
Alexey Dosovitskiy, Lucas Beyer, Alexander Kolesnikov, Dirk Weissenborn, Xiaohua Zhai, Thomas Unterthiner, Mostafa Dehghani, Matthias Minderer, Georg Heigold, Sylvain Gelly, et~al.
\newblock An image is worth 16x16 words: Transformers for image recognition at scale.
\newblock \emph{arXiv preprint arXiv:2010.11929}, 2020.

\bibitem[Epstein et~al.(2017)Epstein, Patai, Julian, and Spiers]{epstein2017cognitive}
Russell~A Epstein, Eva~Zita Patai, Joshua~B Julian, and Hugo~J Spiers.
\newblock The cognitive map in humans: spatial navigation and beyond.
\newblock \emph{Nature neuroscience}, 20\penalty0 (11):\penalty0 1504--1513, 2017.

\bibitem[Ericson and Jensfelt(2024)]{ericson2024beyond}
Ludvig Ericson and Patric Jensfelt.
\newblock Beyond the frontier: Predicting unseen walls from occupancy grids by learning from floor plans.
\newblock \emph{IEEE Robotics and Automation Letters}, 2024.

\bibitem[Gadre et~al.(2023)Gadre, Wortsman, Ilharco, Schmidt, and Song]{gadre2023cows}
Samir~Yitzhak Gadre, Mitchell Wortsman, Gabriel Ilharco, Ludwig Schmidt, and Shuran Song.
\newblock Cows on pasture: Baselines and benchmarks for language-driven zero-shot object navigation.
\newblock In \emph{Proceedings of the IEEE/CVF Conference on Computer Vision and Pattern Recognition}, pages 23171--23181, 2023.

\bibitem[Garg et~al.(2024)Garg, Rana, Hosseinzadeh, Mares, S{\"u}nderhauf, Dayoub, and Reid]{garg2024robohop}
Sourav Garg, Krishan Rana, Mehdi Hosseinzadeh, Lachlan Mares, Niko S{\"u}nderhauf, Feras Dayoub, and Ian Reid.
\newblock Robohop: Segment-based topological map representation for open-world visual navigation.
\newblock \emph{arXiv preprint arXiv:2405.05792}, 2024.

\bibitem[Georgakis et~al.(2022{\natexlab{a}})Georgakis, Bucher, Schmeckpeper, Singh, and Daniilidis]{georgakis2021learning}
Georgios Georgakis, Bernadette Bucher, Karl Schmeckpeper, Siddharth Singh, and Kostas Daniilidis.
\newblock Learning to map for active semantic goal navigation.
\newblock \emph{International Conference on Learning Representations}, 2022{\natexlab{a}}.

\bibitem[Georgakis et~al.(2022{\natexlab{b}})Georgakis, Schmeckpeper, Wanchoo, Dan, Miltsakaki, Roth, and Daniilidis]{georgakis2022cross}
Georgios Georgakis, Karl Schmeckpeper, Karan Wanchoo, Soham Dan, Eleni Miltsakaki, Dan Roth, and Kostas Daniilidis.
\newblock Cross-modal map learning for vision and language navigation.
\newblock In \emph{Proceedings of the IEEE/CVF conference on computer vision and pattern recognition}, pages 15460--15470, 2022{\natexlab{b}}.

\bibitem[Gervet et~al.(2023)Gervet, Chintala, Batra, Malik, and Chaplot]{navigating_real_world}
Theophile Gervet, Soumith Chintala, Dhruv Batra, Jitendra Malik, and Devendra~Singh Chaplot.
\newblock Navigating to objects in the real world.
\newblock \emph{Science Robotics}, 8\penalty0 (79):\penalty0 eadf6991, 2023.

\bibitem[Gireesh et~al.(2022)Gireesh, Kiran, Banerjee, Sridharan, Bhowmick, and Krishna]{gireesh2022object}
Nandiraju Gireesh, DA~Sasi Kiran, Snehasis Banerjee, Mohan Sridharan, Brojeshwar Bhowmick, and Madhava Krishna.
\newblock Object goal navigation using data regularized q-learning.
\newblock In \emph{2022 IEEE 18th International Conference on Automation Science and Engineering (CASE)}, pages 1092--1097. IEEE, 2022.

\bibitem[Gu et~al.(2023)Gu, Kuwajerwala, Morin, Jatavallabhula, Sen, Agarwal, Rivera, Paul, Ellis, Chellappa, et~al.]{gu2023conceptgraphs}
Qiao Gu, Alihusein Kuwajerwala, Sacha Morin, Krishna~Murthy Jatavallabhula, Bipasha Sen, Aditya Agarwal, Corban Rivera, William Paul, Kirsty Ellis, Rama Chellappa, et~al.
\newblock Conceptgraphs: Open-vocabulary 3d scene graphs for perception and planning.
\newblock \emph{arXiv preprint arXiv:2309.16650}, 2023.

\bibitem[He et~al.(2022)He, Chen, Xie, Li, Doll{\'a}r, and Girshick]{he2022masked}
Kaiming He, Xinlei Chen, Saining Xie, Yanghao Li, Piotr Doll{\'a}r, and Ross Girshick.
\newblock Masked autoencoders are scalable vision learners.
\newblock In \emph{Proceedings of the IEEE/CVF conference on computer vision and pattern recognition}, pages 16000--16009, 2022.

\bibitem[Huang et~al.(2023)Huang, Mees, Zeng, and Burgard]{huang2023visual}
Chenguang Huang, Oier Mees, Andy Zeng, and Wolfram Burgard.
\newblock Visual language maps for robot navigation.
\newblock In \emph{2023 IEEE International Conference on Robotics and Automation (ICRA)}, pages 10608--10615. IEEE, 2023.

\bibitem[Jatavallabhula et~al.(2023)Jatavallabhula, Kuwajerwala, Gu, Omama, Chen, Maalouf, Li, Iyer, Saryazdi, Keetha, et~al.]{jatavallabhula2023conceptfusion}
Krishna~Murthy Jatavallabhula, Alihusein Kuwajerwala, Qiao Gu, Mohd Omama, Tao Chen, Alaa Maalouf, Shuang Li, Ganesh Iyer, Soroush Saryazdi, Nikhil Keetha, et~al.
\newblock Conceptfusion: Open-set multimodal 3d mapping.
\newblock \emph{arXiv preprint arXiv:2302.07241}, 2023.

\bibitem[Katyal et~al.(2019)Katyal, Popek, Paxton, Burlina, and Hager]{katyal2019uncertainty}
Kapil Katyal, Katie Popek, Chris Paxton, Phil Burlina, and Gregory~D Hager.
\newblock Uncertainty-aware occupancy map prediction using generative networks for robot navigation.
\newblock In \emph{2019 International Conference on Robotics and Automation (ICRA)}, pages 5453--5459. IEEE, 2019.

\bibitem[Li et~al.(2022)Li, Weinberger, Belongie, Koltun, and Ranftl]{lseg}
Boyi Li, Kilian~Q Weinberger, Serge Belongie, Vladlen Koltun, and Ren{\'e} Ranftl.
\newblock Language-driven semantic segmentation.
\newblock \emph{arXiv preprint arXiv:2201.03546}, 2022.

\bibitem[Li et~al.(2023{\natexlab{a}})Li, Li, Savarese, and Hoi]{li2023blip}
Junnan Li, Dongxu Li, Silvio Savarese, and Steven Hoi.
\newblock Blip-2: Bootstrapping language-image pre-training with frozen image encoders and large language models.
\newblock In \emph{International conference on machine learning}, pages 19730--19742. PMLR, 2023{\natexlab{a}}.

\bibitem[Li et~al.(2023{\natexlab{b}})Li, Debnath, Stein, and Ko{\v{s}}eck{\'a}]{li2023learning}
Yimeng Li, Arnab Debnath, Gregory~J Stein, and Jana Ko{\v{s}}eck{\'a}.
\newblock Learning-augmented model-based planning for visual exploration.
\newblock In \emph{2023 IEEE/RSJ International Conference on Intelligent Robots and Systems (IROS)}, pages 5165--5171. IEEE, 2023{\natexlab{b}}.

\bibitem[Mishra et~al.(2020)Mishra, Garg, Narang, and Mishra]{mishra2020drone}
Balmukund Mishra, Deepak Garg, Pratik Narang, and Vipul Mishra.
\newblock Drone-surveillance for search and rescue in natural disaster.
\newblock \emph{Computer Communications}, 156:\penalty0 1--10, 2020.

\bibitem[Qu et~al.(2024)Qu, Tan, Zhang, Xia, Cadena, and Hutter]{qu2024ippon}
Kaixian Qu, Jie Tan, Tingnan Zhang, Fei Xia, Cesar Cadena, and Marco Hutter.
\newblock Ippon: Common sense guided informative path planning for object goal navigation.
\newblock \emph{3rd Workshop on Language and Robot Learning: Language as an Interface}, 2024.

\bibitem[Radford et~al.(2021{\natexlab{a}})Radford, Kim, Hallacy, Ramesh, Goh, Agarwal, Sastry, Askell, Mishkin, Clark, et~al.]{clip}
Alec Radford, Jong~Wook Kim, Chris Hallacy, Aditya Ramesh, Gabriel Goh, Sandhini Agarwal, Girish Sastry, Amanda Askell, Pamela Mishkin, Jack Clark, et~al.
\newblock Learning transferable visual models from natural language supervision.
\newblock In \emph{International conference on machine learning}, pages 8748--8763. PMLR, 2021{\natexlab{a}}.

\bibitem[Radford et~al.(2021{\natexlab{b}})Radford, Kim, Hallacy, Ramesh, Goh, Agarwal, Sastry, Askell, Mishkin, Clark, et~al.]{radford2021learning}
Alec Radford, Jong~Wook Kim, Chris Hallacy, Aditya Ramesh, Gabriel Goh, Sandhini Agarwal, Girish Sastry, Amanda Askell, Pamela Mishkin, Jack Clark, et~al.
\newblock Learning transferable visual models from natural language supervision.
\newblock In \emph{International conference on machine learning}, pages 8748--8763. PmLR, 2021{\natexlab{b}}.

\bibitem[Ramrakhya et~al.(2023)Ramrakhya, Batra, Wijmans, and Das]{ramrakhya2023pirlnav}
Ram Ramrakhya, Dhruv Batra, Erik Wijmans, and Abhishek Das.
\newblock Pirlnav: Pretraining with imitation and rl finetuning for objectnav.
\newblock In \emph{Proceedings of the IEEE/CVF Conference on Computer Vision and Pattern Recognition}, pages 17896--17906, 2023.

\bibitem[Rana et~al.(2023)Rana, Haviland, Garg, Abou-Chakra, Reid, and Suenderhauf]{rana2023sayplan}
Krishan Rana, Jesse Haviland, Sourav Garg, Jad Abou-Chakra, Ian Reid, and Niko Suenderhauf.
\newblock Sayplan: Grounding large language models using 3d scene graphs for scalable task planning.
\newblock In \emph{7th Annual Conference on Robot Learning}, 2023.

\bibitem[Ronneberger et~al.(2015)Ronneberger, Fischer, and Brox]{unet}
Olaf Ronneberger, Philipp Fischer, and Thomas Brox.
\newblock U-net: Convolutional networks for biomedical image segmentation.
\newblock In \emph{Medical image computing and computer-assisted intervention--MICCAI 2015: 18th international conference, Munich, Germany, October 5-9, 2015, proceedings, part III 18}, pages 234--241. Springer, 2015.

\bibitem[Schmid et~al.(2022)Schmid, Cheema, Reijgwart, Siegwart, Tombari, and Cadena]{schmid2022scexplorer}
Lukas Schmid, Mansoor~Nasir Cheema, Victor Reijgwart, Roland Siegwart, Federico Tombari, and Cesar Cadena.
\newblock Sc-explorer: Incremental 3d scene completion for safe and efficient exploration mapping and planning.
\newblock \emph{arXiv preprint arXiv:2208.08307}, 2022.

\bibitem[Vobecky et~al.(2023)Vobecky, Sim\'{e}oni, Hurych, Gidaris, Bursuc, P\'{e}rez, and Sivic]{pop3d}
Antonin Vobecky, Oriane Sim\'{e}oni, David Hurych, Spyridon Gidaris, Andrei Bursuc, Patrick P\'{e}rez, and Josef Sivic.
\newblock Pop-3d: Open-vocabulary 3d occupancy prediction from images.
\newblock In \emph{Advances in Neural Information Processing Systems}, pages 50545--50557. Curran Associates, Inc., 2023.

\bibitem[Xing et~al.(2023)Xing, Cioffi, Hidalgo-Carri{\'o}, and Scaramuzza]{xing2023autonomous}
Jiaxu Xing, Giovanni Cioffi, Javier Hidalgo-Carri{\'o}, and Davide Scaramuzza.
\newblock Autonomous power line inspection with drones via perception-aware mpc.
\newblock In \emph{2023 IEEE/RSJ International Conference on Intelligent Robots and Systems (IROS)}, pages 1086--1093. IEEE, 2023.

\bibitem[Yamauchi(1997)]{yamauchi1997frontier}
Brian Yamauchi.
\newblock A frontier-based approach for autonomous exploration.
\newblock In \emph{Proceedings 1997 IEEE International Symposium on Computational Intelligence in Robotics and Automation CIRA'97.'Towards New Computational Principles for Robotics and Automation'}, pages 146--151. IEEE, 1997.

\bibitem[Ye et~al.(2021)Ye, Batra, Das, and Wijmans]{ye2021auxiliary}
Joel Ye, Dhruv Batra, Abhishek Das, and Erik Wijmans.
\newblock Auxiliary tasks and exploration enable objectnav.
\newblock \emph{arXiv preprint arXiv:2104.04112}, 2021.

\bibitem[Yokoyama et~al.(2023)Yokoyama, Ha, Batra, Wang, and Bucher]{vlfm}
Naoki~Harrison Yokoyama, Sehoon Ha, Dhruv Batra, Jiuguang Wang, and Bernadette Bucher.
\newblock Vlfm: Vision-language frontier maps for zero-shot semantic navigation.
\newblock In \emph{2nd Workshop on Language and Robot Learning: Language as Grounding}, 2023.

\bibitem[Zeng et~al.(2024)Zeng, Zhang, Ehsani, Hendrix, Salvador, Herrasti, Girshick, Kembhavi, and Weihs]{zeng2024poliformer}
Kuo-Hao Zeng, Zichen Zhang, Kiana Ehsani, Rose Hendrix, Jordi Salvador, Alvaro Herrasti, Ross Girshick, Aniruddha Kembhavi, and Luca Weihs.
\newblock Poliformer: Scaling on-policy rl with transformers results in masterful navigators.
\newblock \emph{Conference on Robot Learning}, 2024.

\bibitem[Zhang et~al.(2024{\natexlab{a}})Zhang, Qu, Patil, Cadena, and Hutter]{zhang2024tag}
Mike Zhang, Kaixian Qu, Vaishakh Patil, Cesar Cadena, and Marco Hutter.
\newblock Tag map: A text-based map for spatial reasoning and navigation with large language models.
\newblock \emph{Conference on Robot Learning}, 2024{\natexlab{a}}.

\bibitem[Zhang et~al.(2024{\natexlab{b}})Zhang, Yu, Song, Wang, and Jiang]{zhang2024imagine}
Sixian Zhang, Xinyao Yu, Xinhang Song, Xiaohan Wang, and Shuqiang Jiang.
\newblock Imagine before go: Self-supervised generative map for object goal navigation.
\newblock In \emph{Proceedings of the IEEE/CVF Conference on Computer Vision and Pattern Recognition}, pages 16414--16425, 2024{\natexlab{b}}.

\bibitem[Zheng et~al.(2020)Zheng, Zhang, Li, Tang, Gao, and Zhou]{zheng2020structured3d}
Jia Zheng, Junfei Zhang, Jing Li, Rui Tang, Shenghua Gao, and Zihan Zhou.
\newblock Structured3d: A large photo-realistic dataset for structured 3d modeling.
\newblock In \emph{Computer Vision--ECCV 2020: 16th European Conference, Glasgow, UK, August 23--28, 2020, Proceedings, Part IX 16}, pages 519--535. Springer, 2020.

\bibitem[Zhou et~al.(2023)Zhou, Zheng, Pryor, Shen, Jin, Getoor, and Wang]{zhou2023esc}
Kaiwen Zhou, Kaizhi Zheng, Connor Pryor, Yilin Shen, Hongxia Jin, Lise Getoor, and Xin~Eric Wang.
\newblock Esc: Exploration with soft commonsense constraints for zero-shot object navigation.
\newblock In \emph{International Conference on Machine Learning}, pages 42829--42842. PMLR, 2023.

\end{thebibliography}
